\documentclass[letterpaper, 10 pt, conference]{ieeeconf}  

\IEEEoverridecommandlockouts                              

\usepackage{graphics} 
\usepackage{epsfig} 
\usepackage{times} 
\usepackage{amsmath} 
\usepackage{amssymb}  
\usepackage{xcolor}
\usepackage[ruled,vlined,linesnumbered]{algorithm2e}
\usepackage{multirow}
\usepackage[a-2b,mathxmp]{pdfx}
\usepackage{pgffor}
\usepackage{tabularx}

\newcolumntype{Y}{>{\centering\arraybackslash}X}

\newcommand{\revise}[1]{#1}
\newcommand{\revisef}[1]{#1}

\usepackage{color, soul} 

\newcommand{\fig}[1]{Fig.~\ref{#1}}  
\newcommand{\tbl}[1]{Table~\ref{#1}} 
\newcommand{\algo}[1]{Algorithm~\ref{#1}} 
\newcommand{\sect}[1]{Sec.~\ref{#1}} 
\newcommand{\etal}[0]{{\em et al.~}}

\usepackage{hyperref}
\usepackage{xspace}
\newcommand{\zephyr}[0]{ZePHyR\xspace}

\title{\LARGE \bf
OSSID: Online Self-Supervised Instance Detection \\by (and for) Pose Estimation
}

\author{Qiao Gu, Brian Okorn, and David Held%
\thanks{Q. Gu is with Department of Computer Science, University of Toronto. ({\tt\small qgu@cs.toronto.edu})}%
\thanks{B. Okorn and D. Held are with the Robotics Institute, Carnegie Mellon University. ({\tt\small bokorn@andrew.cmu.edu, dheld@andrew.cmu.edu})}%
}

\begin{document}

\maketitle
\thispagestyle{empty}
\pagestyle{empty}


\begin{abstract}


Real-time object pose estimation is necessary for many robot manipulation algorithms. However, state-of-the-art methods for object pose estimation are trained for a specific set of objects; these methods thus need to be retrained to estimate the pose of each new object, often requiring tens of GPU-days of training for optimal performance. In this paper, we propose the OSSID framework, leveraging a slow zero-shot pose estimator to self-supervise the training of a fast detection algorithm. This fast detector can then be used to filter the input to the pose estimator, drastically improving its inference speed. We show that this self-supervised training exceeds the performance of existing zero-shot detection methods on two widely used object pose estimation and detection datasets, without requiring any human annotations. Further, we show that the resulting method for pose estimation has a significantly faster inference speed, due to the ability to filter out large parts of the image.  Thus, our method for self-supervised online learning of a detector (trained using pseudo-labels from a slow pose estimator) leads to accurate pose estimation at real-time speeds, without requiring human annotations. Supplementary materials and code can be found at \href{https://georgegu1997.github.io/OSSID/}{https://georgegu1997.github.io/OSSID/} 



\end{abstract} 

\section{Introduction}


Object instance detection and pose estimation are crucial to many robot manipulation tasks. Unlike the standard computer vision tasks of detecting all instances of a given semantic object category (such as person, car, or bicycle), for robotic manipulation, robots  need to  detect specific object instances. For example, a robot agent in the kitchen needs to distinguish a salt can from a coffee can, even though both objects may fall into the semantic category of ``can."

In the past few years, deep convolutional neural networks have become the prevailing tool for object instance detection and pose estimation, outperforming alternative methodologies in various benchmarks~\cite{hodan2019photorealistic,xiang2017posecnn,Li2019-cdpn,labbe2020-cosypose,hodan2018bop}.
However, most deep object detection and pose estimation methods are object-specific, which means the object categories are pre-defined and ``baked into" the network weights. In other words, when we want to apply the detector and pose estimator to new objects, new data needs to be collected and the network needs to be retrained or finetuned on those new object instances. 
Current state-of-the-art methods for object pose estimation~\cite{labbe2020-cosypose} require tens of GPU-days of training, which is quite a long time for a robot to wait every time a human selects a new brand of coffee. This quickly becomes infeasible as every new can, box, and tool requires us to repeat this process.

\begin{figure}[t!]
    \centering
    \includegraphics[width=\linewidth,trim={1cm 0 1cm 0},clip]{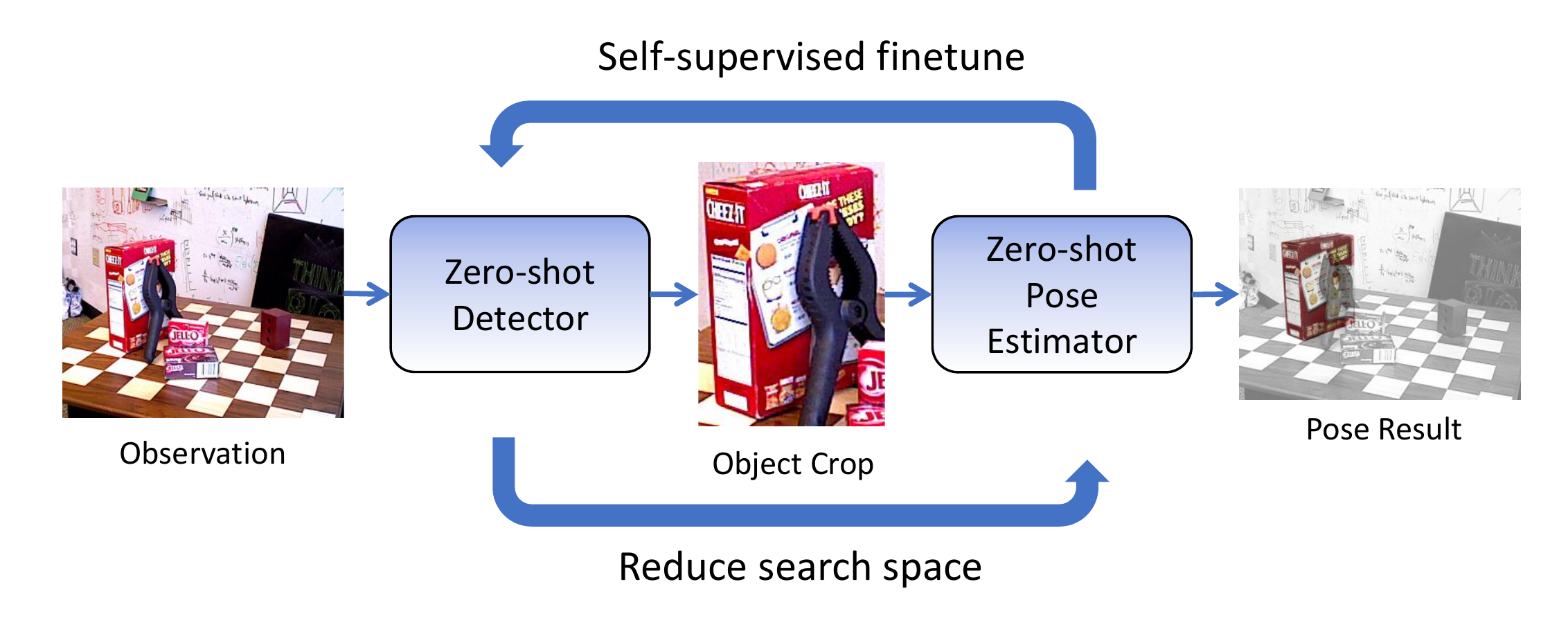}
    \caption{We propose \revisef{OSSID}, a self-supervised learning pipeline for object instance detection by pose estimation. The results of a zero-shot pose estimation network are used to finetune a zero-shot detector online. Then the detection results in turn provide object bounding boxes and reduce the search space for pose estimation. Without any manual annotation required, both the detector and the pose estimator get better and faster. }
    \label{fig:teaser}
\end{figure}

To address this issue, a number of zero-shot pose estimators have been developed. However, most zero-shot pose estimators only evaluate on sparse, uncluttered scenes where the object of interest is detected and cropped or is sitting on an empty table~\cite{xiao2019pose, park2019latentfusion, sundermeyer2020multipath}. Evaluation of such methods in cluttered settings shows that such methods fail to provide reasonable performance, even with the addition of ground-truth bounding boxes or ground truth translation as input (see analysis in Okorn, \etal\cite{icra2021zephyr} Appendix B).
A recent method has directly tackled the challenge of zero-shot object pose estimation in clutter~\cite{icra2021zephyr}, but this approach has a very slow inference speed of 3 seconds per frame, due to the need to generate and evaluate pose hypotheses over a large 6D pose search space.  
This inference time is much too slow for real-time robotics applications.

In this work, we explore how a zero-shot object detector can be combined with a zero-shot pose estimator for faster performance, without loss of accuracy. Specifically, we build upon the work of Okorn~\etal\cite{icra2021zephyr}, but significantly increase the inference speed by using a zero-shot object detector to reduce the search space. This focuses the pose estimation on the smaller region of the image within a detected bounding box, instead of processing the entire image. 

Unfortunately, a naive implementation of this straightforward combination does not give satisfactory performance. As our analysis shows, current zero-shot instance detectors only have mediocre performance when evaluated on objects outside their training set. Therefore, we propose to adapt the detector to novel objects and unseen environments with a zero-shot pose estimator. 






We exploit the insight that a slow method for zero-shot pose estimation provides free and high-quality pseudo-ground truth for training a fast object detector.
As outlined in \fig{fig:teaser}, we propose \revisef{OSSID, an online self-supervised instance detection framework,} using a zero-shot pose estimation pipeline to generate pseudo-ground truth detection and segmentation labels on the test environment. 
After performing self-supervised online learning, the resulting method for object instance detection and pose estimation outperforms baselines on both speed and accuracy by a large margin. 
In this work, we assume that the 3D mesh model of the target object is available; 3D object mesh models can be easily obtained using 3D reconstruction software~\cite{Weise2008-accurateandrobust, zhou2013dense, Weise2011-onlineloop, Krainin2010-manipulatorandobject, Wang2019-inhandobjectscanning}, with an overhead of only several minutes. 

We evaluate \revisef{OSSID} on two popular and challenging datasets: LineMOD-Occlusion and YCB-Video. The results demonstrate that online self-supervised learning using a zero-shot pose estimator can help a detector quickly adapt to new objects and new environments. Further, the detector reduces the search space for 6D poses and drastically improves the inference speed for pose estimation. 


\section{Related Work}

\subsection{Zero-shot Pose Estimation}
Classical methods based on hand-crafted features perform no learning and thus are inherently zero-shot to different object instances. 
In the field of 6D pose estimation, Point Pair Features (PPF) and its variants can still achieve good results on recent benchmarks~\cite{drost2010ppf, drost20123d, kim2011object, hinterstoisser2016going, vidal2018method, hodan2018bop}. However, PPF-based methods have much slower inference speed than deep-learned methods (typically an order of magnitude slower) and no longer match the accuracy of recent deep-learned methods. 

Although deep learning methods for 6D pose estimation have achieved very accurate results, most such methods are trained for particular objects and do not generalize to unseen objects without retraining, which can take tens of GPU-days~\cite{Brachmann2014learning, xiang2017posecnn, wang2019densefusion, labbe2020-cosypose}. Several recent works tackled the zero-shot pose estimation problem by learning a latent object representation~\cite{xiao2019pose, park2019latentfusion, sundermeyer2020multipath}. 
However, recent analysis has revealed that such methods perform poorly in cluttered scenes, even when a ground-truth bounding box is provided as input~\cite{icra2021zephyr}.
In recent work, \zephyr~\cite{icra2021zephyr} overcomes this issue by learning to score many pose hypotheses in cluttered scenes.  However, 
\zephyr requires scoring a large number of pose hypotheses over the entire image.  This results in a slow inference speed of up to 3 seconds per image per object, which prevents its use for real-time robotics applications. We propose to use self-supervised online learning to train a zero-shot instance detector to filter the input to \zephyr, significantly speeding up its performance, without requiring any human annotations.

Related to our work, previous work has shown that the
accuracy and inference speed of PPF can be improved when augmented with a deep learned object instance detection algorithm~\cite{konig2020hybrid}. 
This work, however, required the training of an object-specific pose estimator, which itself requires large training times. In contrast, our approach can be trained quickly with online self-supervised learning.


\subsection{Few-shot Object Detection}
Much effort has been devoted to zero-shot or few-shot object detection by the computer vision community in the past few years~\cite{Kang2019-fewshotobject, Wang2020-frustratinglysimple, Wang2020-democraticattention, Boudiaf2020-fewshotseg, Fan2020-fewshotobject}, but most of these works have been focused on class-level semantic object detection. 
However, in the context of robot manipulation, robot agents often need to locate a specific object instance in a cluttered environment. 
Traditional methods tackled this problem using hand-crafted features and template matching~\cite{Lowe2004-distinctiveimagefeatures, Fergus2003-objectclassrecognition, collet2011moped, Tang2012-texturedobjectrecognition}.  
Recently, several deep learned methods have been proposed for few-shot object instance detection~\cite{ammiratoTDID18,Mercier2020-dtoid}. 
While we build on the state-of-the-art zero-shot instance detector, DTOID~\cite{Mercier2020-dtoid}, we demonstrate that this network only achieves mediocre performance on unseen object instances. We then show that our self-supervised learning pipeline can significantly improve the detection performance. We additionally show we can achieve a faster inference speed with a reduced number of object templates, while maintaining equivalent or better detection accuracy. 

\subsection{Domain Adaptation for Object Detection}


Although deep CNNs have achieved significant progress on object detection, their performance will degrade on images out of the training distribution. Recent efforts have attempted to tackle this challenge by unsupervised domain adaptation~\cite{Chen2018-daf,He2019-multiadv,Hsu2020-progressiveda,Xu2020-crossdomaindetection,Vs2021-megacda}. 
We refer readers to Oza, \etal\cite{Oza2021-udasurvey} for a comprehensive survey on unsupervised domain adaptation of object detection. 
Our method is similar to those based on pseudo-label based self-training~\cite{RoyChowdhury2019-automaticadaptation,Khodabandeh2019-arobustlearning,Kim2019-selftraining}, but we focus on object instance detection (rather than class level detection) and we obtain pseudo labels from zero-shot pose estimation. 
In the related area of domain adaptation for object tracking,
Pirk~\etal\cite{Pirk2019-onlineobject} uses a contrastive loss 
to learn object representations,
showing that tracking performance increases with gradual online training. 
More closely related to our work, Mitash~\etal\cite{Mitash2017-selfsupervisedlearning} proposed a self-supervised online learning system for object detection using physics simulation and multi-view pose estimation. However, this method relies on large synthetic datasets, and their system does not generalize to objects that do not exist in the synthetic training set. In contrast, our method is able to adapt to unseen objects, improving the initial zero-shot detector as more scenes of the test environment are processed. Mitash~\etal also assume the environment to be a clean tabletop or predefined shelves, limiting their range of application. In contrast, our method has been shown to work in cluttered environments, and is able to adapt to objects and environments outside of those it was initially trained in.

\section{Method}
Our goal is to train a fast and accurate pose estimator without requiring any human annotations or long training times.
To achieve this, we \revisef{proposed OSSID: online self-supervised instance detection}, using a slow zero-shot pose estimator (ZePHyR~\cite{icra2021zephyr}) to train a fast object instance detector through online self-supervision. This object instance detector can then be used to filter the input space of our pose estimator, increasing the inference speed without reducing the accuracy of the overall system. 

\subsection{Zero-shot Pose Estimation}

\begin{figure}
    \centering
    \includegraphics[width=1.0\linewidth]{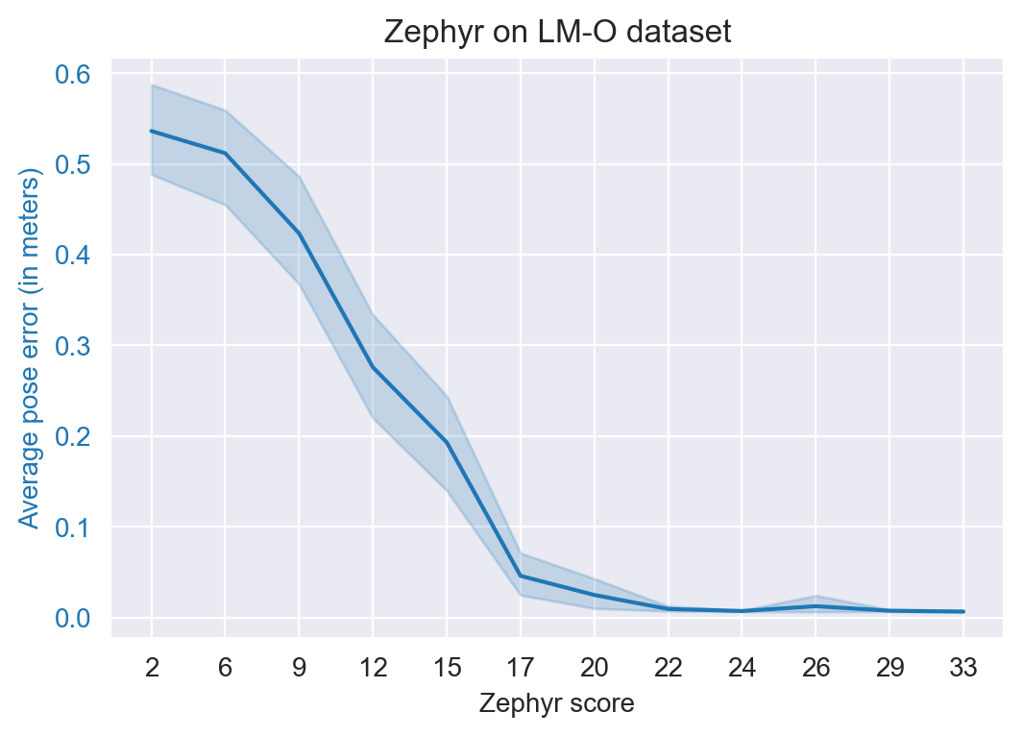}
    \caption{Distribution of mean pose accuracy vs. Zephyr score of the top-scored pose hypothesis for all images in LM-O dataset, with 2 Std Dev shown in light blue. 
    The plot demonstrates that \zephyr returns higher scores for pose hypotheses with lower error. We use pose results with scores higher than 20 for self-supervised learning. }
    \label{fig:zephyr-score}
\end{figure}

Recent work has shown that combining non-learned pose hypothesis generation with a deep learned fitness function can produce highly accurate pose estimates on objects never seen at training time~\cite{icra2021zephyr}. This method, while having the ability to generalize to arbitrary objects with no retraining required, requires a long inference time to generate potential hypotheses across the full image space. 

For object pose estimation, we adopt ZePHyR~\cite{icra2021zephyr}, a zero-shot pose scoring algorithm that generalizes to unseen objects without needing extra labeling or re-training. 
Following~\cite{icra2021zephyr}, we use Point Pair Features~\cite{drost2010ppf} and SIFT feature matching~\cite{lowe1999object} for 6D object pose hypothesis generation. 



The runtime of this pose estimation algorithm is strongly correlated to the number of pose hypotheses being evaluated. 
Therefore, we propose to use an object detector to filter the pose search space, removing unlikely regions of the input.
Specifically, we crop the input scene using a learned object instance detector, generating hypotheses using only points from within the cropped region.
As such, we do not generate hypotheses outside of this bounding box, which reduces the number of hypotheses that \zephyr~will evaluate, which reduces the inference time.  Also, we do not generate features for the region outside of the detector bounding box, which further reduces the runtime.
The combination of these benefits leads to a significant increase in inference speed, as described in \sect{sec-time}.


\begin{figure*}
    \centering
    \includegraphics[width=\linewidth]{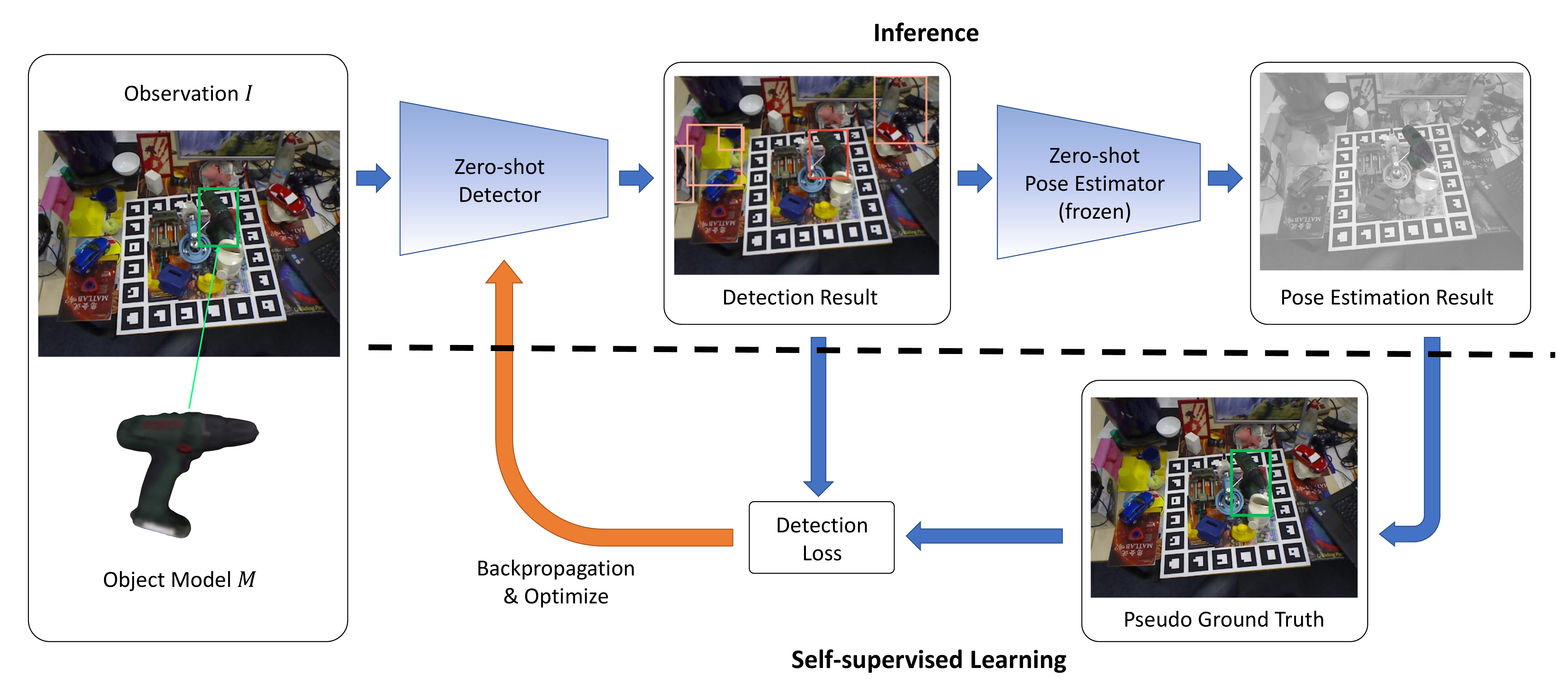}
    \caption{\revisef{Pipeline for OSSID: online self-supervised instance detection by (and for)} 6D pose estimation. \revise{The upper part shows the pipeline for inference, where object detection results are used to filter the input of the pose estimator, increasing its inference speed. The lower part shows the self-supervised learning procedure. High scoring poses generated by the zero-shot pose estimator are used as pseudo ground truth to self-supervise the detection network, which helps the detector adapt and achieve better performance without any manual labels. The weights of the zero-shot pose estimator are not changed in this process. }
    }
    \label{fig:pipeline}
\end{figure*}

\subsection{Zero-shot Detection}

As explained above, we aim to use a detector to filter pose hypotheses for \zephyr, increasing the inference speed of the pose estimation pipeline. However, most object detectors are trained on a large dataset of the target objects, which requires waiting for many GPU-days for training to complete. In contrast, we hope to obtain a pose estimation system that can work quickly on a novel object, which requires that the detection system that will be used to filter hypotheses must be trained quickly as well.

\revise{Recently, researchers have proposed zero-shot methods for object instance detection (such as DTOID~\cite{Mercier2020-dtoid}). These networks are specially designed to compare an object template with the observation of a scene to find the target object.  DTOID is  trained on a large set of objects in an attempt to generalize this comparison to new objects without extra training. }
\revise{Specifically, for DTOID to detect a new object, it only requires template images of the target object.  These template images can be generated by rendering images of an object mesh model~\cite{Mercier2020-dtoid}. We assume such object meshes are available, since they are provided for us in the datasets that we use for evaluation.  For creating such mesh models,  there are many techniques for 3D object reconstruction~\cite{Weise2008-accurateandrobust, zhou2013dense, Weise2011-onlineloop, Krainin2010-manipulatorandobject, Wang2019-inhandobjectscanning}; furthermore, with the help of 3D capturing software, one can  capture a 3D mesh model of the object with a cellphone within several minutes~\cite{software-itseez3d, software-qlone, software-trnio}.}

However, we find that DTOID only has mediocre performance when tested on objects outside of the training distribution, as we show in our experiments in \sect{sec-onlineself}. 
We hypothesize that this is because of the large domain gap between the objects and environments during training and testing. Without adaptation to the unseen testing domain, the network generalizes poorly to out-of-distribution test examples.




To overcome this limitation, we propose a method for online self-supervised finetuning for object instance detection. Specifically, we evaluate a zero-shot pose estimator on previous frames of the target environment. We then use these pose estimates as pseudo-labels for self-supervision.  Given these pseudo-labels, we can finetune an object instance detector, improving its performance on the target environment. The integration of this self-supervised detector will then improve the speed of the overall system by filtering hypotheses for the pose estimator. 




\subsection{Online Self-supervised Learning}


We introduce \revisef{OSSID,} a self-supervised learning framework for online adaptation of an object detector to novel objects in an unseen testing environment, as depicted in \fig{fig:pipeline}. The key insight is that a 6D pose estimator predicts the full state of a rigid object; a zero-shot pose estimator can thus provide free supervision for training an object detector. As the pose estimator is only used for training, even a slow pose estimation method can be used, as speed is less relevant at training time.
Once self-supervised finetuning is complete, the object detector provides a target region for the pose estimator and thus improves both the speed and accuracy of the pose estimator. 

We assume access to a stream of RGB-D images $I_t$, each containing a target object \revise{with its mesh model $M$ and the template images $T$}; we process these images in sequential order, following the online learning paradigm~\cite{Pirk2019-onlineobject}. 
\revise{Note that in this streaming pipeline, each image is seen only once. The detection and pose estimation results are used for both evaluation and online self-supervised learning. }
This aligns with the situation where a robot agent enters a new environment, encounters a new object, and gradually adapts its detector to find this object. 


To obtain pseudo-ground truth labels for training the detector, we run the zero-shot pose estimation network, \zephyr, on the observation image $I$.  The output of \zephyr is a pose estimation result $\{h, s\}$, where $h$ is a pose and $s$ is the score of the associated hypothesis. The question is: when should we trust a pose hypothesis output by the zero-shot pose estimation network?  We ideally only want to train the object detector on pose hypotheses that are accurate, which we can treat as pseudo-ground truth.

Fortunately, we found that the score $s$ generated by \zephyr~is a strong indicator of the accuracy of the pose hypothesis $h$. 
We visualize this in \fig{fig:zephyr-score}, which clearly demonstrates that, as the \zephyr~score increases, the error of the estimated pose goes down.  As the figure shows, pose estimates with \zephyr~score greater than 20 are extremely accurate, with an average pose error of less than 2 centimeters. We can thus filter out pose hypotheses with a score less than 20; we treat pose hypotheses with a score of at least 20 as pseudo-ground truth for training an object detector.
Note that we only consider the highest-scoring pose hypothesis in a given image, ignoring the possibility of training from multiple potential instances of the same object in a scene.


Thus, if the score $s$ is larger than the threshold, we treat this pose as pseudo-ground truth and push it into a finetuning dataset $F$ together with the observation $I$. 
This finetuning dataset is then used to finetune the detection network \revise{through backpropagation, as shown in Figure~\ref{fig:pipeline} (``Backpropagation and optimize").  Note that the input object templates (the rendered images) are not changed during training.}

Note that 6D pose results can be easily converted to detection bounding boxes and segmentation masks by projecting the object model into the image frame.  Thus, a pose estimate provides full supervision for training the detection network (bounding boxes or segmentation masks) without any human annotations. 

To speed up the training process, rather than running the pose estimator on the entire image, we instead run the detector on the image $I$. 
We then only run the pose estimation method on a cropped region within the highest-scoring bounding box. We perform such cropping during both online self-supervised training of the detector (to speed up training) as well as during inference.

The full online self-supervised learning process is further detailed in~\algo{alg:pipeline}. Note that in this pipeline, the object detector is trained on the test dataset, but no manual annotations are used. As we can see from \sect{sec-progress}, the detection accuracies improve as the online self-supervised learning proceeds. 


\begin{algorithm}
\SetKwComment{Comment}{//}{}
\SetKwFunction{Detect}{Detect}
\SetKwFunction{Pose}{Pose}
\SetKwFunction{Crop}{Crop}
\SetKwFunction{GetBest}{GetBest}
\SetKwFunction{Append}{Append}
\SetKwFunction{Length}{Length}
\SetKwFunction{Finetune}{Finetune}
\SetKwInOut{Input}{input}
\SetKwInOut{Output}{output}

\Input{A testing dataset $D$ containing images $I$, \revise{a target object mesh $M$ with template images $T$}, a detector \Detect, a pose estimator \Pose, score threshold for good pose $\theta^p$ 
}
\Output{A fine-tuned detector \Detect}

\caption{OSSID: Online Self-supervised Instance Detection}
\label{alg:pipeline}
\SetAlgoLined

$F \gets \{\}$ \Comment*{Dataset for online learning}
\ForEach{$(I, M)$ in $D$}{
 $d  \gets$ \Detect{$I, \revise{T}$}\;
 $I' \gets$  \Crop{$I$, $d$}\;
  $\{h, s\} \gets$ \Pose{$I', M$}\;
  \If{$s > \theta^p$}{
    \Append{$F$, $(I, h)$}\;
  }
    \Finetune{\Detect, $F$}\;
}
\Return \Detect
\end{algorithm}

\section{Experiments}

\begin{table*}[th!]
\center
\begin{tabular}{c|c|ccccc|c|c}
\hline
Dataset                & Task & DTOID & ZePHyR & \revisef{OSSID (Ours)} & \begin{tabular}[c]{@{}c@{}}\revisef{OSSID}\\ (w/ Conf. Filter)\end{tabular} & \begin{tabular}[c]{@{}c@{}}\revisef{OSSID}\\ \revise{(Oracle)}\end{tabular} & \begin{tabular}[c]{@{}c@{}}\revisef{OSSID}\\ (Transductive)\end{tabular} &
\begin{tabular}[c]{@{}c@{}}\revisef{CosyPose}~\cite{labbe2020-cosypose}\\ (not Zero-shot)\end{tabular} \\ \hline
\multirow{3}{*}{LM-O}  & Detection mAP & 51.3  & \revise{67.1} & 64.0 & \textbf{67.3} & \revise{74.4} & 78.4 & \revise{90.5} \\
                       & Pose AR score & -- & 59.8 & 61.7 & \textbf{63.9} & \revise{66.8} & 66.0 & \revise{63.3} \\
                       & Inference Time (ms) & 430   & 2949   & \textbf{210} & 710 & \revise{210} & 210 & \revise{69} \\ \hline
\multirow{3}{*}{YCB-V} & Detection mAP &  11.6* & \revise{58.1} & 63.2 & \textbf{64.0} & \revise{79.2} & 68.9 & \revise{86.1} \\
                       & Pose AR Score       & -- & 51.6 & 53.3 & \textbf{55.3} & \revise{55.3} & 57.1 & \revise{57.4} \\
                       & Inference Time (ms) & 430 & 619 & \textbf{320} & 350 & \revise{320} & 320 & \revise{69} \\ \hline
\end{tabular}
\caption{Zero-shot detection and pose estimation results. 
For OSSID, the inference time reported is the total time of running both the detector and the pose estimator, and the detection stage in our methods only takes 50 ms. \revise{We also report the detection mAP of baseline \zephyr by converting its pose estimation results to bounding boxes.} *Note that for the DTOID baseline on YCB-V, we use the weights trained on our own dataset, since the pre-trained model was trained on YCB-V over a long training period (we only compare to methods with short, or non-existent (zero-shot), training times). }
\label{tbl:main_results}
\end{table*}

\subsection{Dataset}

For our experiments, we evaluate over two popular datasets, LM-O and YCB-V, which contain rigid objects with 6D pose annotations. These two datasets are challenging for object detection and pose estimation due to the presence of clutter, occlusions, and lighting variations. In our experiments, we assume the object mesh models are available, but we do not need any manual annotation or synthetic data generation for self-supervised learning. 

\textbf{LineMOD-Occlusion dataset} (LM-O)~\cite{Brachmann2014learning} contains a single scene from the testing set of the larger LineMOD (LM) dataset~\cite{Hinterstoisser2012modelbased}. While LM only provides 6D pose annotation for one object in each scene, LM-O densely annotates all 8 low-textured objects in the selected scene. 
For zero-shot pose estimation, we adopted the \zephyr model~\cite{icra2021zephyr} trained on a synthetic dataset containing LM objects that are not in the LM-O dataset~\cite{hodan2019photorealistic}. 
For detection, we used the DTOID model weights provided by~\cite{Mercier2020-dtoid}, which were also trained on a synthetic dataset containing various objects from BOP datasets, excluding those in LM and LM-O. 
In this way, both the detector and pose estimator did not see the LM-O objects during training and their results are zero-shot. 

\textbf{YCB-Video dataset} (YCB-V)~\cite{xiang2017posecnn} includes 92 RGB-D videos captured with 21 YCB objects~\cite{calli2015ycb}, densely annotated with detection bounding boxes, segmentation masks and 6D poses. This dataset is challenging for pose estimation as the videos have different lighting conditions, occlusions, and sensor noise. 
We evaluate our method on the BOP testing set~\cite{hodan2018bop}, which is a subset of the 12 testing videos originally defined in~\cite{xiang2017posecnn} and contains testing images with higher-quality ground truth poses. 
For zero-shot pose estimation, we followed the testing protocol in Okorn~\etal\cite{icra2021zephyr}, adopting two models trained on complementary object sets and testing them on the objects that were not seen during training. 
For detection, since the model weights provided by Mercier~\etal\cite{Mercier2020-dtoid} use the YCB objects during training, we trained our own DTOID weights by creating a synthetic dataset without objects in YCB-V. 
Although this retrained version of DTOID has poor detection results on the YCB-V dataset at first (detection mAP of 11.6 in \tbl{tbl:main_results}), 
we observed a large improvement in detection after online self-supervised learning, even surpassing the non-zero-shot baseline using the weights provided by Mercier~\etal\cite{Mercier2020-dtoid} (Detection mAP of 63.7). 

\subsection{Metrics}

We evaluate the performance of the detector and the pose estimator before and after self-supervised learning. To evaluate the detector, following previous work~\cite{Mercier2020-dtoid}, we report the detection mean average precision (mAP)~\cite{Everingham2010-pascalvoc} using an IoU threshold of 0.5. For pose estimation accuracy, we follow the BOP Challenge~\cite{hodan2018bop} and report the average recall scores (AR). AR is the average of three pose accuracy metrics: Visible Surface Discrepancy (VSD), Maximum Symmetry-Aware Surface Distance (MSSD), and Maximum Symmetry-Aware Projection Distance (MSPD). For detailed formulations, we refer readers to Hoda\v{n}~\etal\cite{hodan2018bop}.



\subsection{Baselines}
While there are many methods for object detection and pose estimation, most of them require large training datasets and do not generalize to unseen object instances. In contrast, our goal is to design a system for detection and pose estimation that can be trained very quickly without large training datasets.  Therefore, we select DTOID~\cite{Mercier2020-dtoid} and \zephyr~\cite{icra2021zephyr} as the baseline methods for object detection and pose estimation respectively. DTOID~\cite{Mercier2020-dtoid} achieves the state-of-the-art results on zero-shot object instance detection, outperforming all other comparable approaches on the LM-O dataset~\cite{Hinterstoisser2011-multimodaltemplates,ammiratoTDID18,Wang2019-fastonline}. Similarly, \zephyr~\cite{icra2021zephyr} is the state-of-the-art in zero-shot pose estimation on LM-O and YCB-V.


\subsection{Online Self-supervised Learning}
\label{sec-onlineself}
We compare the performance of the detection and pose estimation results of our method to zero-shot baselines in~\tbl{tbl:main_results}.
``\revisef{OSSID (Ours)}'' shows the performance of our method when the instance detector is self-supervised finetuned online following \algo{alg:pipeline}. 
As can be seen in this table, our method outperforms both DTOID and \zephyr on both the detection mAP and pose AR metrics for both the LM-O and YCB-V datasets. Importantly, for pose estimation, we reduce the inference time compared to Zephyr by a factor of 14 on LM-O and a factor of almost 2 on YCB-V.

In our experiments, \revise{the DTOID network is optimized with a fixed set of object templates rendered from the object mesh model. We also experimented with using only 10 templates instead of 160 (though in each experiment, the number of templates is still held fixed throughout training).
We report the difference in \tbl{tab:localtemp}.} We found that, for our method where the network is learning online, reducing the number of templates has little impact on the performance of the detection network, but greatly increased the inference speed. Specifically, we can see that the detector trained using our method only needs 10 local templates to have comparable performance while achieving a real-time speed. \revise{In contrast, the original DTOID network has a large drop in performance on the segmentation mean IOU metric when the number of templates is reduced.} 

\begin{table}[]
    \centering
    \begin{tabular}{c|cccc}
    \hline
    Method                 & \begin{tabular}[c]{@{}c@{}}Number of \\ local templates\end{tabular} & \begin{tabular}[c]{@{}c@{}}Detection\\ Time (ms)\end{tabular} & \begin{tabular}[c]{@{}c@{}}Detection\\ mAP\end{tabular} & \begin{tabular}[c]{@{}c@{}}Segmentation\\ mean IoU\end{tabular} \\ \hline
    \multirow{2}{*}{DTOID} & 10                                                                   & \textbf{50}                                                       & 50.5                                                    & 33.4                                                            \\
                           & 160                                                                  & 430                                                      & 51.3                                                    & 41.6                                                            \\ \hline
    \multirow{2}{*}{\revisef{OSSID}}  & 10                                                                   & \textbf{50}                                                       & \textbf{64.0}                                                    & 46.8                                                            \\
                           & 160                                                                  & 430                                                      & 62.7                                                    & \textbf{48.4}                                                            \\ \hline
    \end{tabular}
    \caption{Effect of local templates on DTOID detection network on LM-O dataset, with and without our self-supervised learning pipeline. }
    \label{tab:localtemp}
\end{table}

We also tested a variant of our method, in which we use an approach that we call ``confidence filtering", reported as ``\revisef{OSSID} (w/ Conf. Filter)'' in \tbl{tbl:main_results}.
In this setting, the input image $I$ will be cropped to the region $I'$ for pose estimation only if the detected bounding boxes have high predicted confidence, as defined by a detection score above a given threshold. 
Otherwise, if the detection score is below the defined threshold, pose estimation will be done using the full image $I$, resulting in a longer processing time.
This confidence filtering is helpful in the early stage of online learning, where the detector may still have poor performance; if we don't use confidence filtering in such cases, the pose estimator will still run on the cropped image from the bounding box of a poor detector,  which may cause the pose estimator to focus on the wrong region of the image. Confidence filtering will reject such low-confidence detections, and instead run the pose estimator on the entire observation. This leads to slower inference speed on average but better performance, as we can see in \tbl{tbl:main_results}. 

\revise{In addition, we further study the potential negative effects of low-accuracy pose estimates in the  pseudo ground truth used in the self-supervised training process, as these bad pose estimates may mislead the detector. We conduct an experiment where the ground truth object pose is used to finetune the detector. Specifically, we modify the online self-supervised protocol to use the ground truth bounding box and mask, instead of the estimated pose $h$, to finetune the detection network (line 7 of~\algo{alg:pipeline}). The results are reported in the ``\revisef{OSSID} (Oracle)'' column in~\tbl{tbl:main_results}. The difference between ``\revisef{OSSID (Ours)}'' and ``\revisef{OSSID} (Oracle)'' shows there is still a gap between using the pseudo and real ground truth. However, labeling such high-quality ground truth requires extensive human efforts; in contrast, our self-supervised learning pipeline demonstrates a way of improving detection without manual labeling.}

\revise{To evaluate the gap that still exists between the zero-shot and object-specific methods, we also report the performance of CosyPose~\cite{labbe2020-cosypose} in \tbl{tbl:main_results} from the ``CosyPose-ECCV20-PBR-1VIEW'' variant on the BOP leaderboard\footnote{https://bop.felk.cvut.cz/leaderboards/}. This method is a state-of-the-art non-zero-shot pose estimator that is trained solely on synthetic data. However, CosyPose requires tens of GPU-days for synthetic data generation and network training. As shown in \tbl{tbl:main_results}, CosyPose (non-zero-shot, trained for tens of GPU-days on a much larger dataset), achieved better results than our approach in detection mAP and similar results to our approach in terms of pose AR score.
However, such methods require either large-scale manual data annotation or synthetic data generation to work on new objects, with long data generation and training times, whereas our method can quickly adapt online to new objects. Further, our method can train directly on real data in a self-supervised way, without requiring manual annotations or synthetic dataset generation.}


\begin{figure}
    \centering
    \includegraphics[width=\linewidth]{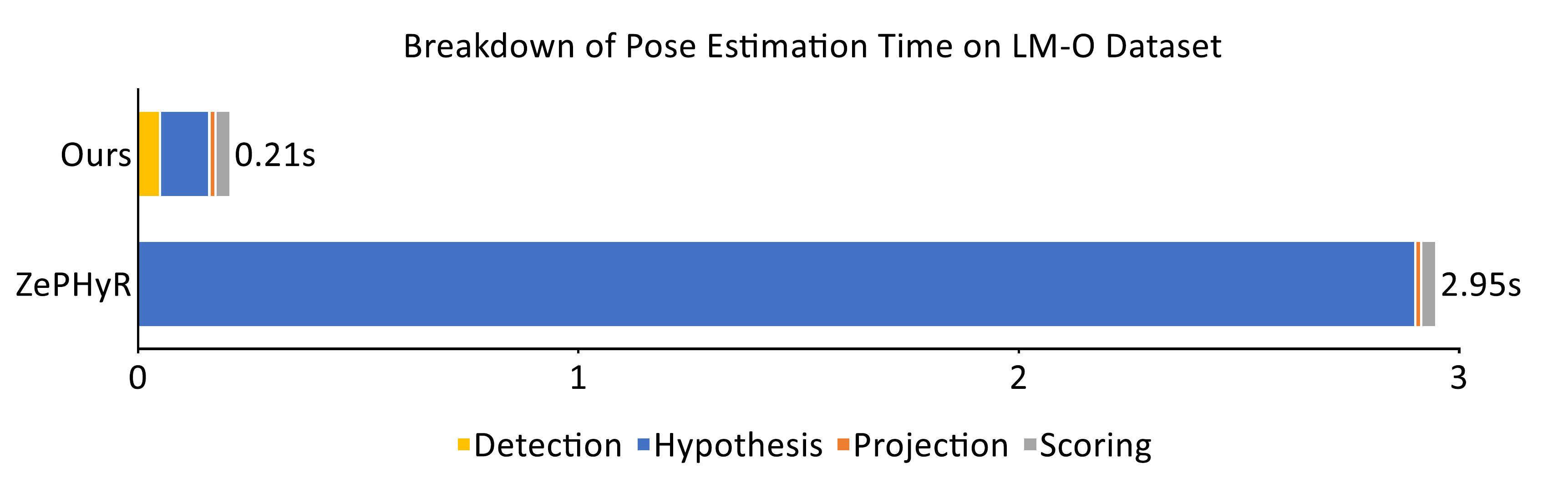}
    \caption{Breakdown of the inference time of pose estimation using \zephyr~\cite{icra2021zephyr} on the LM-O dataset. Note that the time for pose hypothesis generation is greatly reduced on detected regions for our method. 
    }
    \label{fig:time}
\end{figure}

\subsection{Transductive Learning}
\label{sec:exp-transductive}

The performance can be further improved if it is possible to obtain all testing images beforehand and train the network offline with self-supervision. In such scenarios we devise a transductive learning pipeline~\cite{Arnold2007-transductive}, 
where the detector is self-supervised trained and then tested on the same set of testing images (without any annotations). In the offline training stage, the zero-shot pose estimator runs on all testing images first and then the zero-shot detector is self-supervised trained on the pose estimation results (for 50 epochs in our experiments). This offline training process takes less than half an hour on the LM-O and YCB-V BOP testing set. 
The result of this transductive learning pipeline is shown as ``\revisef{OSSID} (Transductive)'' in \tbl{tbl:main_results}; we can see that transductive learning leads to another significant performance boost compared to the online learning pipeline on both datasets. Although this setup may not be used for real-time pose estimation, it can be used for estimating the poses of objects from a fixed dataset. Further, it provides an upper-bound performance of our method, allowing our method to learn from both past and future frames (instead of only learning from past frames). 






\subsection{Time Analysis}
\label{sec-time}

The significant speedup for zero-shot pose estimation results from shrinking the pose search space from the full observation image to just the region within the highest-scoring bounding box of the detector.
We show the inference time breakdown of the pose estimator on the LM-O dataset in \fig{fig:time}. We can see that, although a detection overhead of 50 milliseconds is added compared to \zephyr, the time spent on pose hypothesis generation is reduced by a factor of 25. 
The online finetuning in total takes 6 minutes for the LM-O dataset and about one hour and a half for the YCB-V dataset on a single GPU, which is much less than tens of GPU-days needed to train object-specific detectors. The gradient updates can be run in a background thread and thus would not delay the inference time. 



\begin{figure}
    \centering
    \includegraphics[width=\linewidth]{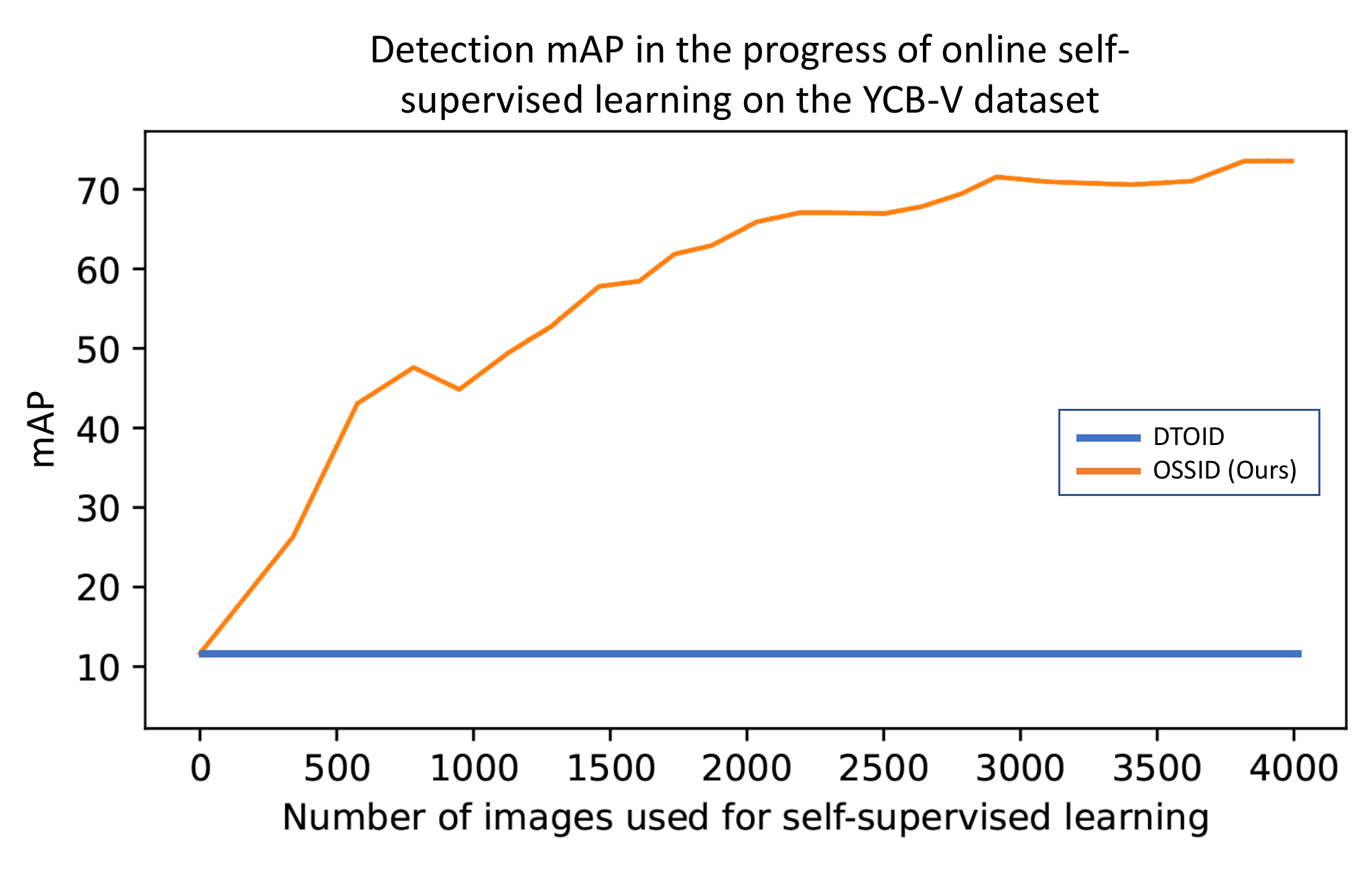}
    \caption{Detection performance improves as more unlabeled test images are used for online self-supervised learning. The OSSID detection models are taken from different timestamps in online learning on the YCB-V dataset and the detection mAP are evaluated on the entire testing set. The detection performance improves significant over time using the proposed OSSID pipeline while the baseline DTOID remains constant performance. }
    \label{fig:progress}
\end{figure}

\subsection{Online Learning}
\label{sec-progress}

To further quantify how the model improves over the course of self-supervised training, we evaluate the model at different points of the OSSID pipeline. 
In \fig{fig:progress}, we take the model weights at different timestamps in self-supervised learning and evaluate them on the entire testing set. 
It can be seen that as the network are self-supervised trained on more and more images of the scene, the accuracy of the detection increases. 
The results demonstrate that, through self-supervision, the instance detector gradually adapts to new objects and environments using the new observations, without the need for annotations. This opens up future directions of applying this method to robot manipulation tasks where the perception system needs to adapt to novel environments. 




\section{Conclusions}

We propose \revisef{a novel method, OSSID,} using a slow zero-shot pose estimator to train a fast detection algorithm, without the need for any annotations. We show that a detector, trained in this self-supervised manner, shows adaptation to new objects and new environments, and exceeds the accuracy of similar zero-shot methods on  cluttered environments. Further, we show that this detector can be used to filter the search space of a zero-shot pose estimator.  This drastically reduces the inference time of the pose estimation system, while maintaining state-of-the-art accuracy. Our method thus shows the benefit of online self-supervised learning, resulting in a high-performance real-time pose estimation system that can be trained within 6 minutes (for the LM-O dataset). 

\section{Acknowledgements}

This work was supported in part by the National Science Foundation Smart and Autonomous Systems Program under Grant IIS-1849154 and in part by LG Electronics.

\section{Appendix}

\subsection{Implementation Details}

In this section, we will provide more details about the implementation of the algorithms described in Sec. III. 

\subsubsection{Online Self-supervised Learning}

The online self-supervised learning pipeline is described in Sec. III-C and Algorithm I. Here we will provide more details for the algorithm. The \texttt{Finetune()} function is called after every 32 finetuning examples of $(I, h)$ are added to the finetuning set $F$.  Within each \texttt{Finetune()}, the detection network is finetuned on the all examples in $F$ for a single epoch. Similar to~\cite{Mercier2020-dtoid}, we use AMS-Grad~\cite{Sashank2018-amsgrad} for optimization with a learning rate of $10^{-4}$, a weight decay of $10^{-6}$ and a batch size of 8. 

\subsubsection{Transductive Learning}
\label{transductive}

The transductive learning pipeline simulates the scenarios where all the testing images are known before the network makes any inference and thus the network can be self-supervised trained on the testing images with more time budgets. Specifically, we run the zero-shot pose estimator on the uncropped images in the testing dataset first and use the pose estimates to train the detector in a regular epoch training fashion. In this way the pose estimation stage is the same as \zephyr~\cite{icra2021zephyr} and we can simply take the results from~\cite{icra2021zephyr} as pseudo ground truth in the implementation. For training, we initialize the weights of the detector as described in Sec. IV-A and finetune it for 50 epochs on the pseudo ground truth. 
We use the same optimizer as in the online learning pipeline, and shrink the learning rate to its tenth at the epoch 20 and 40. 

Here we further report the training time using the transductive learning protocol. 
Since the network is only trained on the testing dataset, which contains 1445 images for LM-O and 4123 images for YCB-V. Therefore the network training for transductive learning only takes 50 epochs is roughly 25 minutes for LM-O and 72 minutes for YCB-V. This demonstrates that the proposed pipeline is capable to adapt to new environments and novel objects in a short time. 

\subsection{Synthetic Data Generation and Training}

Since YCB-V objects were already used for training in~\cite{Mercier2020-dtoid}, we need to re-train the DTOID network using another dataset in order to show the generalization ability to novel objects. Therefore we generated a synthetic dataset using BlenderProc~\cite{Denninger2019-blenderproc}. We adopted the objects from BOP datasets~\cite{hodan2018bop} except for those from LM, LM-O and YCB-V. and additionally used 200 ShapeNet objects~\cite{Chang2015-shapenet} with randomized CC0 textures~\cite{AmbientCG_undated-wi}. 
The scenes were generated by randomly dropping objects onto a table and images were captured at randomly sampled camera poses. In this way, we produced a dataset of 40,000 images and trained the detection network from random weights for 100 epochs. We used the same optimizer and scheduler as described in \sect{transductive}. 

Note that our DTOID weights did not reproduce the zero-shot detection performance as reported in~\cite{Mercier2020-dtoid}. However, the performance of our DTOID model quickly adapts to YCB-V objects as shown in Sec. IV-D and Table I. 

\subsection{Comparison to Non Zero-shot Detector}
\revise{To analyze the benefit of zero-shot networks in our pipeline, we tested our self-supervised learning framework where the DTOID detection network is replaced by a a non zero-shot detector, specifically Mask R-CNN~\cite{he2017mask} with the ResNet-50~\cite{he2016resnet} backbone pretrained on MS COCO dataset~\cite{Lin2015-mscoco}. 
The results on the LM-O dataset are shown in \tbl{tab:maskrcnn}. We found that in the transductive learning setting, Mask R-CNN can yield similar pose estimation performance as DTOID, but worse detection results. 
In the online learning setting, DTOID shows much better performance than Mask R-CNN. The reason might be that we need a much larger dataset to train a non zero-shot object detector, while zero-shot detectors like DTOID are designed to quickly adapt to new objects.}

\setcounter{table}{2}

\begin{table}[]
    \centering
    \begin{tabular}{c|ccc}
    \hline
    Method                                                                           & Task          & Ours & \begin{tabular}[c]{@{}c@{}}Ours \\ w/ Mask R-CNN\end{tabular} \\ \hline
    \multirow{2}{*}{\begin{tabular}[c]{@{}c@{}}Online \\ Learning\end{tabular}}      & Detection mAP & 64.0 & 36.5                                                          \\
                                                                                     & Pose AR score & 61.7 & 43.9                                                          \\ \hline
    \multirow{2}{*}{\begin{tabular}[c]{@{}c@{}}Transductive\\ Learning\end{tabular}} & Detection mAP & 78.4 & 56.1                                                          \\
                                                                                     & Pose AR score & 66.0 & 62.7                                                          \\ \hline
    \end{tabular}
    \caption{Comparison of our results and an ablation  replacing DTOID with Mask R-CNN.  Results are reported on the LM-O dataset.}
    \label{tab:maskrcnn}
\end{table}

\subsection{Qualitative Results}

\begin{figure*}[p]
    \centering
    \includegraphics[width=\linewidth]{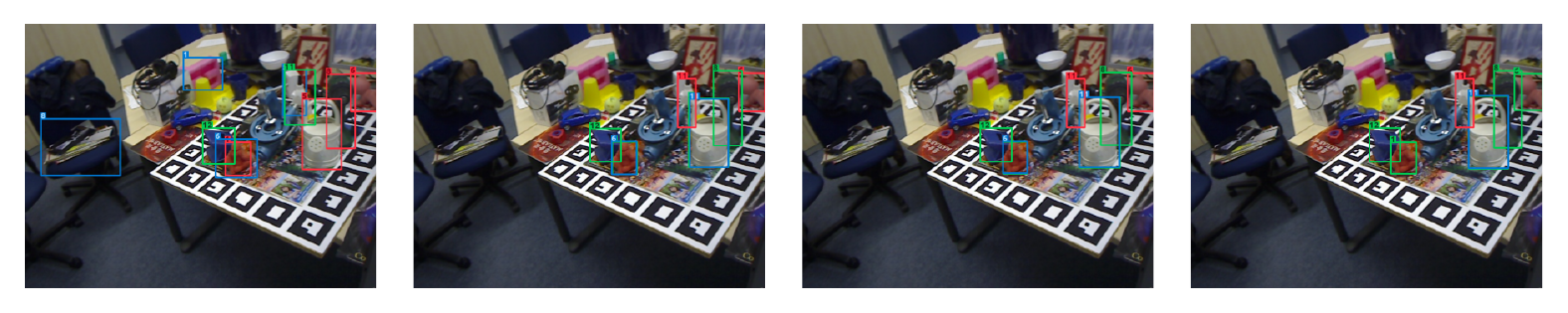}
    \includegraphics[width=\linewidth]{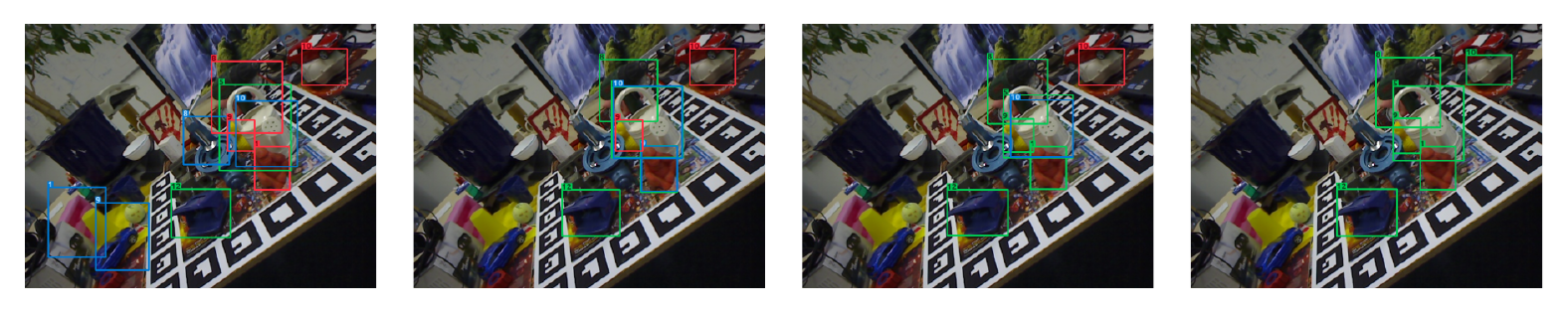}
    \includegraphics[width=\linewidth]{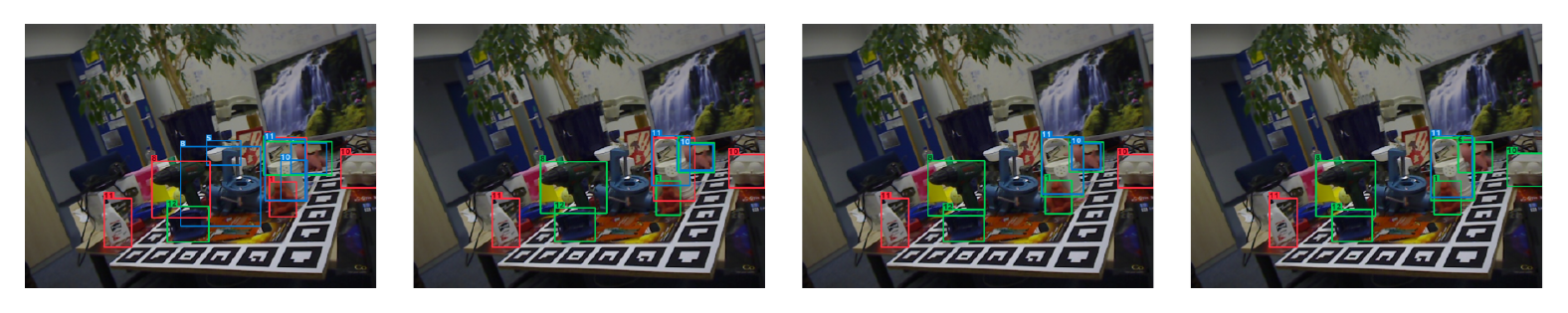}
    \includegraphics[width=\linewidth]{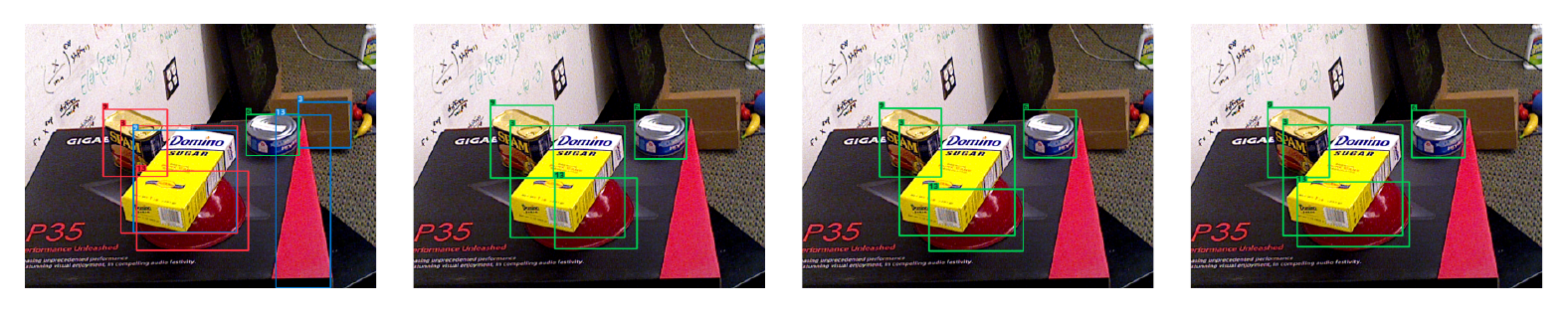}
    \includegraphics[width=\linewidth]{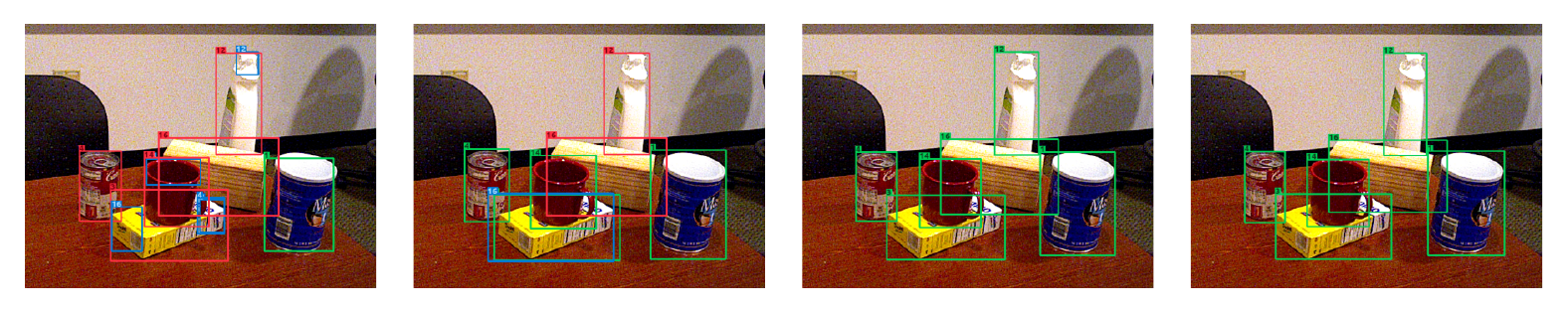}
    \includegraphics[width=\linewidth]{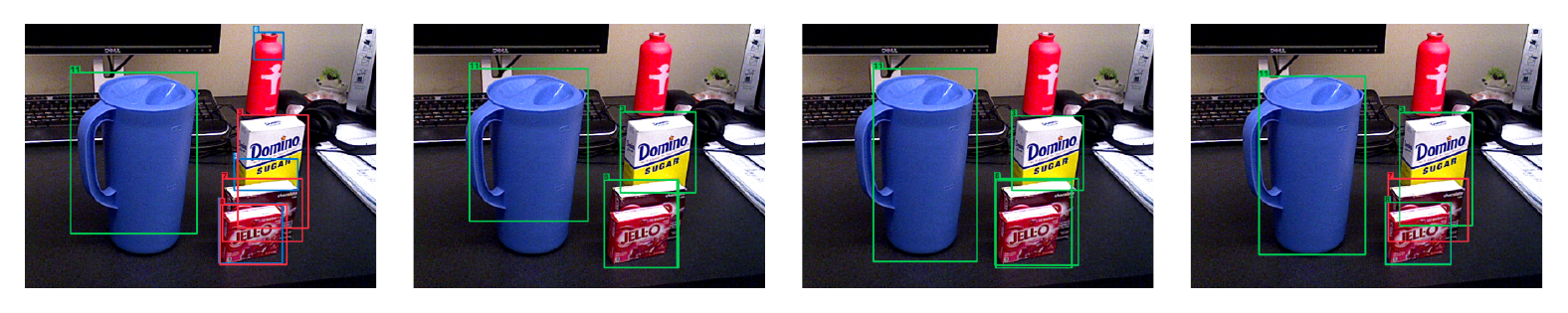}
    \begin{tabularx}{\textwidth}{YYYY}
        (a) Initial model & (b) After 1/3 images & (c) After 2/3 images & (d) After all images\\
    \end{tabularx}
    \caption{Qualitative results of the online self-supervised learned detector, after seeing different number of images. Green, blue and red boxes mean correct, false and missed detection results respectively. The leftmost column shows the results of the baseline detection model and the others show the performance after the detector has been trained on different portion of the test set. The first three rows are the results from the LM-O dataset and the last three rows are from the YCB-V dataset. }
    \label{fig:qual_prog}
\end{figure*}

In \fig{fig:qual_prog}, we show some qualitative results of the detector during the progress of the online self-supervised learning pipeline. Here we recorded the model weights after it is trained on different portion of the test dataset and compare their performance. We can see that the performance gradually improves and the previously missed or false detection are corrected as the online self-supervised learning continues. 

\bibliographystyle{IEEEtran}
\bibliography{IEEEabrv,root}

\begin{thebibliography}{10}
\providecommand{\url}[1]{#1}
\csname url@rmstyle\endcsname
\providecommand{\newblock}{\relax}
\providecommand{\bibinfo}[2]{#2}
\providecommand\BIBentrySTDinterwordspacing{\spaceskip=0pt\relax}
\providecommand\BIBentryALTinterwordstretchfactor{4}
\providecommand\BIBentryALTinterwordspacing{\spaceskip=\fontdimen2\font plus
\BIBentryALTinterwordstretchfactor\fontdimen3\font minus
  \fontdimen4\font\relax}
\providecommand\BIBforeignlanguage[2]{{%
\expandafter\ifx\csname l@#1\endcsname\relax
\typeout{** WARNING: IEEEtran.bst: No hyphenation pattern has been}%
\typeout{** loaded for the language `#1'. Using the pattern for}%
\typeout{** the default language instead.}%
\else
\language=\csname l@#1\endcsname
\fi
#2}}

\bibitem{hodan2019photorealistic}
T.~Hoda{\v{n}}, V.~Vineet, R.~Gal, E.~Shalev, J.~Hanzelka, T.~Connell,
  P.~Urbina, S.~Sinha, and B.~Guenter, ``Photorealistic image synthesis for
  object instance detection,'' in \emph{ICIP}, 2019, pp. 66--70.

\bibitem{xiang2017posecnn}
Y.~Xiang, T.~Schmidt, V.~Narayanan, and D.~Fox, ``Posecnn: A convolutional
  neural network for 6d object pose estimation in cluttered scenes,''
  \emph{RSS}, 2018.

\bibitem{Li2019-cdpn}
Z.~Li, G.~Wang, and X.~Ji, ``\BIBforeignlanguage{en}{{CDPN}:
  {Coordinates-Based} disentangled pose network for {Real-Time} {RGB-Based}
  {6-DoF} object pose estimation},'' in \emph{\BIBforeignlanguage{en}{ICCV}},
  2019, pp. 7677--7686.

\bibitem{labbe2020-cosypose}
Y.~{Labbe}, J.~{Carpentier}, M.~{Aubry}, and J.~{Sivic}, ``Cosypose: Consistent
  multi-view multi-object 6d pose estimation,'' in \emph{ECCV}, 2020, pp.
  574--591.

\bibitem{hodan2018bop}
T.~Hoda{\v{n}}, F.~Michel, E.~Brachmann, W.~Kehl, A.~Glent~Buch, D.~Kraft,
  B.~Drost, J.~Vidal, S.~Ihrke, X.~Zabulis, C.~Sahin, F.~Manhardt, F.~Tombari,
  T.-K. Kim, J.~Matas, and C.~Rother, ``{BOP}: Benchmark for {6D} object pose
  estimation,'' \emph{ECCV}, pp. 19--35, 2018.

\bibitem{xiao2019pose}
Y.~Xiao, X.~Qiu, P.~Langlois, M.~Aubry, and R.~Marlet, ``Pose from shape: Deep
  pose estimation for arbitrary 3d objects,'' in \emph{BMVC}, 2019, p.~61.

\bibitem{park2019latentfusion}
K.~Park, A.~Mousavian, Y.~Xiang, and D.~Fox, ``Latentfusion: End-to-end
  differentiable reconstruction and rendering for unseen object pose
  estimation,'' in \emph{CVPR}, 2020, pp. 10\,707--10\,716.

\bibitem{sundermeyer2020multipath}
M.~Sundermeyer, M.~Durner, E.~Y. Puang, Z.-C. Marton, N.~Vaskevicius, K.~O.
  Arras, and R.~Triebel, ``Multi-path learning for object pose estimation
  across domains,'' in \emph{CVPR}, 2020, pp. 13\,913--13\,922.

\bibitem{icra2021zephyr}
B.~Okorn, Q.~Gu, M.~Hebert, and D.~Held, ``Zephyr: Zero-shot pose hypothesis
  rating,'' in \emph{ICRA}, 2021, pp. 14\,141--14\,148.

\bibitem{Weise2008-accurateandrobust}
T.~Weise, B.~Leibe, and L.~Van~Gool, ``Accurate and robust registration for
  in-hand modeling,'' in \emph{CVPR}, 2008, pp. 1--8.

\bibitem{zhou2013dense}
Q.-Y. Zhou and V.~Koltun, ``Dense scene reconstruction with points of
  interest,'' \emph{ACM ToG}, vol.~32, no.~4, pp. 1--8, 2013.

\bibitem{Weise2011-onlineloop}
T.~Weise, T.~Wismer, B.~Leibe, and L.~Van~Gool, ``Online loop closure for
  real-time interactive {3D} scanning,'' \emph{CVIU}, pp. 635--648, 2011.

\bibitem{Krainin2010-manipulatorandobject}
M.~Krainin, P.~Henry, X.~Ren, and D.~Fox, ``Manipulator and object tracking for
  in hand model acquisition,'' in \emph{ICRA}, 2010, pp. 1817--1824.

\bibitem{Wang2019-inhandobjectscanning}
F.~Wang and K.~Hauser, ``In-hand object scanning via {RGB-D} video
  segmentation,'' in \emph{ICRA}, 2019, pp. 3296--3302.

\bibitem{drost2010ppf}
B.~{Drost}, M.~{Ulrich}, N.~{Navab}, and S.~{Ilic}, ``Model globally, match
  locally: Efficient and robust 3d object recognition,'' in \emph{CVPR}, 2010,
  pp. 998--1005.

\bibitem{drost20123d}
B.~Drost and S.~Ilic, ``3d object detection and localization using multimodal
  point pair features,'' in \emph{3DIMPVT}, 2012, pp. 9--16.

\bibitem{kim2011object}
E.~{Kim} and G.~{Medioni}, ``3d object recognition in range images using
  visibility context,'' in \emph{IROS}, 2011, pp. 3800--3807.

\bibitem{hinterstoisser2016going}
S.~Hinterstoisser, V.~Lepetit, N.~Rajkumar, and K.~Konolige, ``Going further
  with point pair features,'' in \emph{ECCV}, 2016, pp. 834--848.

\bibitem{vidal2018method}
J.~Vidal, C.-Y. Lin, X.~Llad{\'o}, and R.~Mart{\'\i}, ``A method for 6d pose
  estimation of free-form rigid objects using point pair features on range
  data,'' \emph{Sensors}, vol.~18, no.~8, p. 2678, 2018.

\bibitem{Brachmann2014learning}
E.~Brachmann, A.~Krull, F.~Michel, S.~Gumhold, J.~Shotton, and C.~Rother,
  ``Learning 6d object pose estimation using 3d object coordinates,'' in
  \emph{ECCV}, 2014, pp. 536--551.

\bibitem{wang2019densefusion}
C.~Wang, D.~Xu, Y.~Zhu, R.~Mart{\'\i}n-Mart{\'\i}n, C.~Lu, L.~Fei-Fei, and
  S.~Savarese, ``Densefusion: 6d object pose estimation by iterative dense
  fusion,'' in \emph{CVPR}, 2019, pp. 3343--3352.

\bibitem{konig2020hybrid}
R.~K{\"o}nig and B.~Drost, ``A hybrid approach for 6dof pose estimation,'' in
  \emph{ECCV}, 2020, pp. 700--706.

\bibitem{Kang2019-fewshotobject}
B.~Kang, Z.~Liu, X.~Wang, F.~Yu, J.~Feng, and T.~Darrell,
  ``\BIBforeignlanguage{en}{{Few-Shot} object detection via feature
  reweighting},'' in \emph{\BIBforeignlanguage{en}{ICCV}}, 2019, pp.
  8419--8428.

\bibitem{Wang2020-frustratinglysimple}
X.~Wang, T.~E. Huang, J.~Gonzalez, T.~Darrell, and F.~Yu, ``Frustratingly
  simple few-shot object detection,'' in \emph{ICML}, 2020, pp. 9919--9928.

\bibitem{Wang2020-democraticattention}
H.~Wang, X.~Zhang, Y.~Hu, Y.~Yang, X.~Cao, and X.~Zhen,
  ``\BIBforeignlanguage{en}{{Few-Shot} semantic segmentation with democratic
  attention networks},'' in \emph{\BIBforeignlanguage{en}{ECCV}}, 2020, pp.
  730--746.

\bibitem{Boudiaf2020-fewshotseg}
M.~Boudiaf, H.~Kervadec, I.~M. Ziko, P.~Piantanida, I.~B. Ayed, and J.~Dolz,
  ``Few-shot segmentation without meta-learning: {A} good transductive
  inference is all you need?'' in \emph{CVPR}, 2021, pp. 13\,979--13\,988.

\bibitem{Fan2020-fewshotobject}
Q.~Fan, W.~Zhuo, C.-K. Tang, and Y.-W. Tai, ``Few-shot object detection with
  {attention-RPN} and multi-relation detector,'' in \emph{CVPR}, 2020, pp.
  4013--4022.

\bibitem{Lowe2004-distinctiveimagefeatures}
D.~G. Lowe, ``Distinctive image features from scale-invariant keypoints,''
  \emph{IJCV}, vol.~60, no.~2, pp. 91--110, 2004.

\bibitem{Fergus2003-objectclassrecognition}
R.~Fergus, P.~Perona, and A.~Zisserman, ``Object class recognition by
  unsupervised scale-invariant learning,'' in \emph{CVPR}, 2003, pp. 264--271.

\bibitem{collet2011moped}
A.~Collet, M.~Martinez, and S.~S. Srinivasa, ``The moped framework: Object
  recognition and pose estimation for manipulation,'' \emph{IJRR}, vol.~30,
  no.~10, pp. 1284--1306, 2011.

\bibitem{Tang2012-texturedobjectrecognition}
J.~Tang, S.~Miller, A.~Singh, and P.~Abbeel, ``A textured object recognition
  pipeline for color and depth image data,'' in \emph{ICRA}, 2012, pp.
  3467--3474.

\bibitem{ammiratoTDID18}
P.~Ammirato, C.-Y. Fu, M.~Shvets, J.~Kosecka, and A.~C. Berg,
  ``\BIBforeignlanguage{en}{Target driven instance detection},''
  \emph{\BIBforeignlanguage{en}{arXiv:1803.04610 [cs]}}, Oct. 2019.

\bibitem{Mercier2020-dtoid}
J.-P. Mercier, M.~Garon, P.~Gigu{\`e}re, and J.-F. Lalonde, ``Deep
  template-based object instance detection,'' in \emph{WACV}, 2021, pp.
  1506--1515.

\bibitem{Chen2018-daf}
Y.~Chen, W.~Li, C.~Sakaridis, D.~Dai, and L.~Van~Gool, ``Domain adaptive faster
  r-cnn for object detection in the wild,'' in \emph{CVPR}, 2018, pp.
  3339--3348.

\bibitem{He2019-multiadv}
Z.~He and L.~Zhang, ``Multi-adversarial faster-rcnn for unrestricted object
  detection,'' in \emph{ICCV}, 2019, pp. 6668--6677.

\bibitem{Hsu2020-progressiveda}
H.-K. Hsu, C.-H. Yao, Y.-H. Tsai, W.-C. Hung, H.-Y. Tseng, M.~Singh, and M.-H.
  Yang, ``Progressive domain adaptation for object detection,'' in \emph{WACV},
  2020, pp. 749--757.

\bibitem{Xu2020-crossdomaindetection}
M.~Xu, H.~Wang, B.~Ni, Q.~Tian, and W.~Zhang, ``Cross-domain detection via
  graph-induced prototype alignment,'' in \emph{CVPR}, 2020, pp.
  12\,355--12\,364.

\bibitem{Vs2021-megacda}
V.~Vs, V.~Gupta, P.~Oza, V.~A. Sindagi, and V.~M. Patel, ``{MeGA-CDA}: Memory
  guided attention for {Category-Aware} unsupervised domain adaptive object
  detection,'' in \emph{CVPR}, 2021, pp. 4516--4526.

\bibitem{Oza2021-udasurvey}
P.~Oza, V.~A. Sindagi, V.~S. Vibashan, and V.~M. Patel,
  ``\BIBforeignlanguage{en}{Unsupervised domain adaptation of object detectors:
  A survey},'' \emph{\BIBforeignlanguage{en}{arXiv:2105.13502 [cs]}}, May 2021.

\bibitem{RoyChowdhury2019-automaticadaptation}
A.~RoyChowdhury, P.~Chakrabarty, A.~Singh, S.~Jin, H.~Jiang, L.~Cao, and
  E.~Learned-Miller, ``Automatic adaptation of object detectors to new domains
  using self-training,'' in \emph{CVPR}, 2019, pp. 780--790.

\bibitem{Khodabandeh2019-arobustlearning}
M.~Khodabandeh, A.~Vahdat, M.~Ranjbar, and W.~G. Macready, ``A robust learning
  approach to domain adaptive object detection,'' in \emph{ICCV}, 2019, pp.
  480--490.

\bibitem{Kim2019-selftraining}
S.~Kim, J.~Choi, T.~Kim, and C.~Kim, ``Self-training and adversarial background
  regularization for unsupervised domain adaptive one-stage object detection,''
  in \emph{ICCV}, 2019, pp. 6091--6100.

\bibitem{Pirk2019-onlineobject}
S.~Pirk, M.~Khansari, Y.~Bai, C.~Lynch, and P.~Sermanet, ``Online object
  representations with contrastive learning,'' \emph{arXiv:1906.04312 [cs]},
  June 2019.

\bibitem{Mitash2017-selfsupervisedlearning}
C.~Mitash, K.~E. Bekris, and A.~Boularias, ``A self-supervised learning system
  for object detection using physics simulation and multi-view pose
  estimation,'' in \emph{IROS}, 2017, pp. 545--551.

\bibitem{lowe1999object}
D.~G. Lowe, ``Object recognition from local scale-invariant features,'' in
  \emph{ICCV}, 1999, pp. 1150--1157.

\bibitem{software-itseez3d}
itSeez3D, ``\#1 mobile 3d scanning app for ipad | itseez3d,''
  \url{https://itseez3d.com/scanner.html}, accessed: 2021-11-19.

\bibitem{software-qlone}
E.~V.~T. LTD, ``Qlone, 3d scan any object, anywhere!''
  \url{https://www.qlone.pro/}, accessed: 2021-11-19.

\bibitem{software-trnio}
{Trnio}, ``Trnio 3d scanner,'' \url{https://www.trnio.com/}, accessed:
  2021-11-19.

\bibitem{Hinterstoisser2012modelbased}
S.~Hinterstoisser, V.~Lepetit, S.~Ilic, S.~Holzer, G.~Bradski, K.~Konolige, and
  N.~Navab, ``Model based training, detection and pose estimation of
  texture-less 3d objects in heavily cluttered scenes,'' in \emph{ACCV}, 2013,
  pp. 548--562.

\bibitem{calli2015ycb}
B.~{Calli}, A.~{Singh}, A.~{Walsman}, S.~{Srinivasa}, P.~{Abbeel}, and A.~M.
  {Dollar}, ``The ycb object and model set: Towards common benchmarks for
  manipulation research,'' in \emph{ICAR}, 2015, pp. 510--517.

\bibitem{Everingham2010-pascalvoc}
M.~Everingham, L.~Van~Gool, C.~K.~I. Williams, J.~Winn, and A.~Zisserman,
  ``\BIBforeignlanguage{en}{The pascal visual object classes ({VOC})
  challenge},'' \emph{\BIBforeignlanguage{en}{IJCV}}, vol.~88, no.~2, pp.
  303--338, 2010.

\bibitem{Hinterstoisser2011-multimodaltemplates}
S.~Hinterstoisser, S.~Holzer, C.~Cagniart, S.~Ilic, K.~Konolige, N.~Navab, and
  V.~Lepetit, ``Multimodal templates for real-time detection of texture-less
  objects in heavily cluttered scenes,'' in \emph{ICCV}, 2011, pp. 858--865.

\bibitem{Wang2019-fastonline}
Q.~Wang, L.~Zhang, L.~Bertinetto, W.~Hu, and P.~H.~S. Torr, ``Fast online
  object tracking and segmentation: {A} unifying approach,'' in \emph{CVPR},
  2019, pp. 1328--1338.

\bibitem{Arnold2007-transductive}
A.~Arnold, R.~Nallapati, and W.~W. Cohen, ``A comparative study of methods for
  transductive transfer learning,'' in \emph{ICDM Workshops}, 2007, pp. 77--82.

\bibitem{Sashank2018-amsgrad}
J.~R. Sashank, K.~Satyen, and K.~Sanjiv, ``On the convergence of adam and
  beyond,'' in \emph{ICLR}, vol.~5, 2018, p.~7.

\bibitem{Denninger2019-blenderproc}
M.~Denninger, M.~Sundermeyer, D.~Winkelbauer, Y.~Zidan, D.~Olefir,
  M.~Elbadrawy, A.~Lodhi, and H.~Katam,
  ``\BIBforeignlanguage{en}{{BlenderProc}},''
  \emph{\BIBforeignlanguage{en}{arXiv:1911.01911 [cs]}}, Oct. 2019.

\bibitem{Chang2015-shapenet}
A.~X. Chang, T.~Funkhouser, L.~Guibas, P.~Hanrahan, Q.~Huang, Z.~Li,
  S.~Savarese, M.~Savva, S.~Song, H.~Su, J.~Xiao, L.~Yi, and F.~Yu,
  ``\BIBforeignlanguage{en}{{ShapeNet}: An {Information-Rich} {3D} model
  repository},'' \emph{\BIBforeignlanguage{en}{arXiv:1512.03012 [cs]}}, Dec.
  2015.

\bibitem{AmbientCG_undated-wi}
{ambientCG}, ``{ambientCG} - free public domain {PBR} materials,''
  \url{https://ambientcg.com/}, accessed: 2021-9-20.

\bibitem{he2017mask}
K.~He, G.~Gkioxari, P.~Doll{\'a}r, and R.~Girshick, ``Mask r-cnn,'' in
  \emph{ICCV}, 2017, pp. 2961--2969.

\bibitem{he2016resnet}
K.~He, X.~Zhang, S.~Ren, and J.~Sun, ``Deep residual learning for image
  recognition,'' in \emph{CVPR}, 2016, pp. 770--778.

\bibitem{Lin2015-mscoco}
T.~Lin, M.~Maire, S.~J. Belongie, J.~Hays, P.~Perona, D.~Ramanan,
  P.~Doll{\'{a}}r, and C.~L. Zitnick, ``Microsoft {COCO:} common objects in
  context,'' in \emph{ECCV}, 2014, pp. 740--755.

\end{thebibliography}

\end{document}